\documentclass[runningheads]{llncs}
\usepackage[T1]{fontenc}
\usepackage{graphicx,verbatim}
\usepackage{hyperref}
\hypersetup{colorlinks=true, linkcolor=blue, urlcolor=blue, citecolor=blue}
\usepackage{amssymb}
\usepackage{enumitem}
\usepackage{amsmath}
\usepackage{amssymb}
\usepackage{booktabs}
\usepackage{capt-of}
\usepackage{multirow}

\usepackage{tikz}
\usetikzlibrary{positioning, arrows.meta, fit, backgrounds, shapes.geometric}

\begin{document}
\title{One-Shot Data Selection for Medical Image Classification via Graph Coverage}
\titlerunning{Graph Coverage Selection}

\author{Zahiriddin Rustamov\inst{1,2}\orcidID{0000-0003-4977-1781} \and
Nadia Badawi\inst{1}\orcidID{0009-0004-8795-3521} \and
Rafat Damseh\inst{1}\orcidID{0000-0001-6797-0448} \and
Nazar Zaki\inst{1}\orcidID{0000-0002-6259-9843}}
\authorrunning{Z. Rustamov et al.}
\institute{Department of Computer Science and Software Engineering, United Arab Emirates University, Al Ain, United Arab Emirates \\
    \email{\{700043167, 700047012, rdamseh, Nzaki\}@uaeu.ac.ae}
\and Department of Computer Science, KU Leuven, Leuven, Belgium}
  
\maketitle              % typeset the header of the contribution
\begin{abstract}
Training medical image classifiers on entire datasets is wasteful when annotation budgets are limited: not all samples contribute equally, yet acquiring expert labels is expensive.
Active learning reduces annotation cost through iterative querying, but assumes repeated access to an oracle and requires multiple rounds of model training.
One-shot geometry-based methods such as facility location avoid retraining but operate on pairwise distances that ignore the local structure of the data manifold.

We propose a graph-based one-shot selection method that operates entirely on frozen foundation model embeddings.
Given embeddings from a pretrained encoder, we construct a $k$-nearest neighbor graph over all training samples and derive a two-term coverage kernel from the heat diffusion kernel, capturing both direct and two-hop neighborhood relationships.
Greedy facility location on this kernel selects class-balanced subsets that maximize coverage of the data manifold.
The two-term kernel matches the full spectral heat kernel in selection behavior while reducing computation to sparse matrix operations with a single hyperparameter.

We evaluate on five MedMNIST datasets spanning histopathology, radiology, and microscopy, comparing against both training-dynamics and geometry-based baselines.
Our method achieves the highest balanced accuracy on nine of ten dataset-ratio conditions, with the largest gains on class-imbalanced datasets where global graph construction captures cross-class structure that per-class methods miss, all without any model training during selection.
Code is available at \url{https://github.com/zahiriddin-rustamov/graph-coverage-selection}.


\end{abstract}

\section{Introduction}

Automated classification of medical images supports clinical workflows from cancer grading in histopathology to disease screening in radiology, but training accurate classifiers requires large annotated datasets.
In medical imaging, annotation depends on domain experts whose time is scarce and expensive, and data access is often constrained by institutional policies and patient privacy~\cite{budd2021survey,wang2024alsurvey}.
When annotation budgets are limited, the question is not just how many samples to label, but \emph{which} samples to label.

Active learning addresses this through iterative querying: a model is trained, the most informative unlabeled samples are selected, and an expert annotates them~\cite{settles2009active}.
However, it assumes ongoing access to an annotation oracle and requires multiple rounds of training and querying.
Annotation budgets are often fixed upfront by study protocols or institutional review, and repeated expert access is impractical; even in cold-start settings, no standard method consistently outperforms random sampling~\cite{colossal2023}.
\emph{One-shot} data selection, where the subset is chosen before any training, avoids these constraints.

Outside medical imaging, one-shot selection has been studied through two families of methods.
\emph{Training-dynamics} approaches score samples based on their behavior during training: EL2N~\cite{paul2021el2n} uses early-training error norms, forgetting events~\cite{toneva2019forgetting} count classification reversals across epochs, and EVA~\cite{eva2024} identifies samples with high cross-epoch variance.
These methods are effective~\cite{sorscher2022beyond} but require training a model before any selection occurs, which is circular when the goal is to reduce training cost.
\emph{Geometry-based} approaches operate directly on feature representations: facility location~\cite{wei2015submodularity} selects subsets that maximize coverage in feature space and herding~\cite{welling2009herding} iteratively matches subset statistics to the full dataset.
These methods rely on pairwise distances without modeling local manifold structure: two samples may be equidistant from a third but lie in regions of very different density.

The availability of foundation models pretrained on large-scale medical image collections changes what is feasible for one-shot selection.
Models such as UNI~\cite{chen2024uni} produce high-quality embeddings without task-specific training.
Several methods leverage such embeddings for medical data selection: ENRICH~\cite{enrich2023} applies diversity sampling on learned features and confounder-aware approaches~\cite{ji2025} mitigate confounding variable bias, but both still operate on flat pairwise similarities.
A $k$-nearest neighbor graph on these embeddings captures local manifold structure explicitly: propagation through the graph reveals redundancy and coverage relationships that flat distance matrices miss.
While graphs are well-established in medical image analysis for \emph{prediction}~\cite{zhou2019cgcnet,pati2022hactnet,chen2021patchgcn}, using graphs to model \emph{sample-to-sample} relationships for data selection is, to our knowledge, unexplored in medical imaging.

We propose a graph-based one-shot selection method for medical image classification.
Given frozen UNI embeddings, we build a global $k$-NN graph over all training samples and compute a coverage kernel $\mathbf{K} = \mathbf{A}_{\text{sym}} + \mathbf{A}_{\text{sym}}^2$ that captures both direct and indirect neighborhood relationships.
This kernel arises from decomposition of the heat diffusion kernel; we show that only the first two polynomial terms are necessary, reducing computation to sparse matrix operations.
Greedy facility location on this kernel selects diverse, representative subsets with per-class budget constraints.

Our contributions are:
\begin{enumerate}
    \item A one-shot selection method for medical image classification that operates on frozen foundation model embeddings, requiring no model training before or during selection.
    \item Evidence that graph-based selection requires only local adjacency: a two-term polynomial on the normalized adjacency matches the full spectral heat kernel in selection quality, reducing computation from cubic eigendecomposition to linear-time sparse operations.
    \item Evaluation on five MedMNIST datasets demonstrating that global graph construction across classes captures decision-boundary information that per-class methods miss, with stronger gains on class-imbalanced datasets.
\end{enumerate}

\section{Method}

\begin{figure}[!ht]
\centering
\input{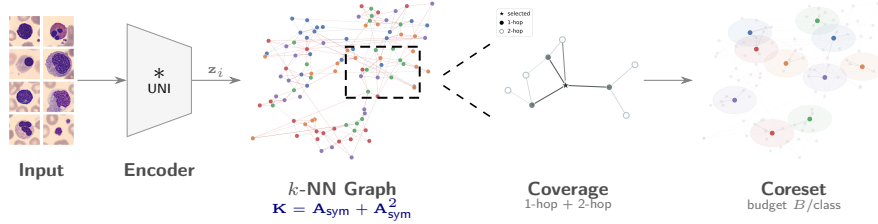}
\caption{Overview of graph coverage selection (illustrated on RetinaMNIST). A frozen UNI encoder maps images to embeddings. A $k$-NN graph is built over all samples, and the coverage kernel $\mathbf{K}$ captures both direct and 2-hop neighborhood structure (center panel: \textbf{$\star$}\,selected, $\bullet$\,1-hop, $\circ$\,2-hop). Greedy facility location selects a balanced coreset maximizing total coverage.}
\label{fig:overview}
\end{figure}

Let $\mathcal{D} = \{(\mathbf{x}_i, y_i)\}_{i=1}^{N}$ be a labeled training set with $C$ classes, and $f: \mathcal{X} \to \mathbb{R}^d$ a pretrained foundation model encoder.
Given a per-class budget $B$, we select a subset $\mathcal{S} \subset \mathcal{D}$ with $|\mathcal{S}_c| = B$ for each class $c$ to maximize downstream classification accuracy.
Selection operates on frozen embeddings $\mathbf{z}_i = f(\mathbf{x}_i)$ (Fig.~\ref{fig:overview}).

\subsection{Neighborhood Graph from Embeddings}

The relationship between two samples depends on their shared context: two samples may be equidistant from a third, but one shares a dense local neighborhood with it while the other is isolated.
A $k$-nearest neighbor graph captures this local context by connecting each sample only to its closest neighbors, preserving the manifold structure of the embedding space.

We construct a $k$-NN graph $\mathcal{G}$ over all $N$ embeddings using cosine similarity.
Embeddings are $L_2$-normalized and each sample is connected to its $k$ nearest neighbors in Euclidean space.
Let $\mathbf{A} \in \mathbb{R}^{N \times N}$ denote the weighted adjacency matrix, with edge weights reflecting pairwise similarity.
We symmetrize: $\mathbf{A} \leftarrow \max(\mathbf{A}, \mathbf{A}^\top)$, so that neighborhood is a mutual relationship.

\subsection{Coverage Kernel via Multi-Hop Propagation}

The adjacency matrix $\mathbf{A}$ encodes direct (1-hop) neighbor relationships.
However, two samples that are not direct neighbors may still be redundant if they share many common neighbors.
Capturing this requires looking beyond direct connections.
We achieve this through a two-step process: symmetric normalization followed by squaring the adjacency.
We first normalize $\mathbf{A}$ to account for varying node degrees:
\begin{equation}
    \mathbf{A}_{\text{sym}} = \mathbf{D}^{-1/2} \mathbf{A} \, \mathbf{D}^{-1/2},
\end{equation}
where $\mathbf{D} = \text{diag}(\mathbf{A} \mathbf{1})$ is the degree matrix.
This symmetric normalization ensures that if sample $i$ covers sample $j$, then $j$ equally covers $i$, which is necessary for consistent coverage accounting during selection.
The coverage kernel is then:
\begin{equation}
    \mathbf{K} = \mathbf{A}_{\text{sym}} + \mathbf{A}_{\text{sym}}^2.
    \label{eq:kernel}
\end{equation}
Entry $K_{ij}$ quantifies how well selecting sample $j$ would cover sample $i$, accounting for both direct proximity and shared neighborhood structure.

\noindent\textbf{Relationship to the heat diffusion kernel.}
The heat diffusion kernel on graphs, $\mathbf{K}_t = \exp(-t\mathbf{L})$ where $\mathbf{L} = \mathbf{I} - \mathbf{A}_{\text{sym}}$ is the normalized Laplacian, is a canonical framework for measuring proximity via diffusion processes on manifolds~\cite{kondor2002diffusion}.
Expanding as a matrix polynomial gives $\mathbf{K}_t = \sum_{k=0}^{\infty} c_k \mathbf{A}_{\text{sym}}^k$ with Poisson-distributed coefficients.
Ablations (Table~\ref{tab:ablation}) show which components of this expansion are necessary for data selection:
\begin{itemize}
    \item Eigendecomposition is unnecessary: the polynomial form suffices.
    \item Terms beyond $\mathbf{A}_{\text{sym}}^2$ contribute negligibly.
    \item Decay coefficients provide no benefit over uniform weighting.
\end{itemize}
The two-term polynomial and full heat kernel achieve comparable accuracy and near-identical per-sample coverage scores ($\rho > 0.999$, Table~\ref{tab:ablation}).
This reduces computation from $O(N^3)$ eigendecomposition to $O(Nk^2)$ sparse matrix multiplication, with $k$ as the sole hyperparameter.

\subsection{Greedy Selection with Per-Class Budget Constraints}

Given the kernel $\mathbf{K}$, we select $B$ samples per class to maximize total coverage using greedy facility location~\cite{nemhauser1978submodular}.
We construct the graph \emph{globally} across all classes rather than independently per class.
Per-class graphs can only optimize within-class coverage.
A global graph also encodes cross-class proximity: samples near the decision boundary have non-zero kernel entries to samples of other classes, making them more valuable for coverage because selecting them simultaneously represents their own class neighborhood and captures information about the class boundary.

Let $\text{cov}_i = \max_{j \in \mathcal{S}} K_{ij}$ denote the current coverage of sample $i$ by the selected set $\mathcal{S}$.
At each step, we select the sample with the largest marginal coverage gain:
\begin{equation}
    j^* = \arg\max_{j \notin \mathcal{S},\; |\mathcal{S}_{y_j}| < B} \sum_{i=1}^{N} \max(0,\; K_{ij} - \text{cov}_i),
    \label{eq:greedy}
\end{equation}
where the constraint $|\mathcal{S}_{y_j}| < B$ enforces that class $y_j$ has not reached its budget.
The facility location objective is monotone submodular under per-class budget constraints (a partition matroid), and greedy maximization provides a $1/2$-approximation guarantee~\cite{fisher1978analysis}.

\section{Experiments}

\subsection{Experimental Setup}

\noindent\textbf{Datasets.}
We evaluate on MedMNIST v2~\cite{yang2023medmnist}, a standardized collection of medical image classification benchmarks.
We select five multi-class datasets spanning multiple modalities and scales:
\textbf{OrganAMNIST} (34,561 training, 11 classes, abdominal CT axial view),
\textbf{OrganSMNIST} (13,932 training, 11 classes, abdominal CT sagittal view),
\textbf{PathMNIST} (89,996 training, 9 classes, colorectal cancer histology),
\textbf{TissueMNIST} (165,466 training, 8 classes, kidney cortex microscopy),
and \textbf{BloodMNIST} (11,959 training, 8 classes, blood cell microscopy).
We use \textbf{DermaMNIST} (7,007 training, 7 classes, dermatoscopy) for the global graph analysis due to its severe class imbalance (58:1 ratio).

\noindent\textbf{Embeddings.}
We use the $224 \times 224$ MedMNIST+ variant and extract embeddings using UNI~\cite{chen2024uni}, a ViT-L/16 pretrained on over 100 million histopathology images.
Images are passed through the frozen encoder, yielding 1024-dimensional embeddings.

\noindent\textbf{Training Protocol.}
For each selected subset, we train a ResNet-18~\cite{he2016resnet} from scratch at $224 \times 224$ resolution using SGD (learning rate 0.1, momentum 0.9, weight decay $5 \times 10^{-4}$) for 1,000 epochs with batch size 256.
We use cosine annealing for learning rate scheduling and no data augmentation, to isolate the effect of data selection from training regularization.

\noindent\textbf{Baselines.}
i) \textbf{Random}: balanced random sampling with equal samples per class.
\emph{Training-based:}
ii) \textbf{EL2N}~\cite{paul2021el2n}: selects by error $L_2$-norm from early training epochs.
iii) \textbf{Forgetting}~\cite{toneva2019forgetting}: selects the most forgotten samples, counting classification reversals across full training.
iv) \textbf{EVA}~\cite{eva2024}: selects samples with high training-error variability across consecutive epochs, requiring 200 epochs of pre-training.
\emph{One-shot geometry:}
v) \textbf{Facility Location}~\cite{guo2022deepcore}: greedy coverage maximization on full pairwise cosine similarity, the standard geometry-based coreset method.
vi) \textbf{FPS}: farthest point sampling in embedding space, selecting for maximum spread.
vii) \textbf{Herding}~\cite{welling2009herding}: iterative mean-matching that selects samples to minimize the distance between subset and full-class centroids in embedding space.
All methods use balanced per-class budgets.

\noindent\textbf{Evaluation.}
We report balanced accuracy.
Each configuration is run with 5 random trials (varying training seed), and we report mean $\pm$ standard deviation.
Selection ratios: 2\% and 5\% of training data.

\subsection{Main Results}

Table~\ref{tab:main} reports balanced accuracy across five datasets at 2\% and 5\% selection ratios.
Our method achieves the highest balanced accuracy on four of five datasets at 2\% and all five at 5\%, with the largest gains on PathMNIST (+3.9 percentage points (pp) over facility location at 2\%) and BloodMNIST (+5.2\,pp at 5\%).
Among one-shot methods, herding is the closest competitor, trailing by small margins on OrganSMNIST ($+$0.5\,pp) and OrganAMNIST ($+$0.2\,pp at 2\%) but falling further behind on PathMNIST and BloodMNIST where local manifold structure is more heterogeneous.
Herding operates per-class by design (matching subset centroids to class centroids) and cannot leverage cross-class graph structure (Table~\ref{tab:global}).
Across all ten dataset-ratio conditions, our method outperforms both herding and facility location on nine (sign test, $p = 0.02$); the sole exception is TissueMNIST at 2\%, where embedding quality bounds all methods.
Facility location, herding, and our method outperform all three training-dynamics baselines on every dataset at both ratios, indicating that at low selection budgets, representation-based coverage is more effective than difficulty-based scoring.
On TissueMNIST, where full-data accuracy remains below 70\%~\cite{yang2023medmnist} due to limited discriminability in UNI embeddings for fluorescence microscopy, the coverage-based methods cluster within 2\,pp of each other, as selection quality is bounded by representation quality.

\begin{table}[t]
\caption{Balanced accuracy (\%) on five MedMNIST datasets. \textbf{Bold}: best, \underline{underline}: second best. Mean$\pm$std over 5 trials.}\label{tab:main}
\centering
\setlength{\tabcolsep}{2.5pt}
\fontsize{8}{9.5}\selectfont
\begin{tabular}{ll cccc cccc}
\toprule
& & & \multicolumn{3}{c}{\textit{Training-dynamics}} & \multicolumn{4}{c}{\textit{One-shot geometry}} \\
\cmidrule(lr){4-6} \cmidrule(lr){7-10}
Dataset & Ratio & Random & EL2N & Forg. & EVA & Facility & FPS & Herding & Ours \\
\midrule
\multirow{2}{*}{OrganS}
 & 2\% & 57.2{\tiny$\pm$4.7} & 39.5{\tiny$\pm$2.0} & 42.5{\tiny$\pm$1.4} & 52.3{\tiny$\pm$1.3} & \underline{63.3}{\tiny$\pm$1.2} & 59.6{\tiny$\pm$2.7} & 63.2{\tiny$\pm$0.6} & \textbf{63.7}{\tiny$\pm$0.7} \\
 & 5\% & 66.0{\tiny$\pm$1.2} & 51.4{\tiny$\pm$0.9} & 56.6{\tiny$\pm$1.1} & 66.0{\tiny$\pm$0.9} & 67.7{\tiny$\pm$1.1} & 66.0{\tiny$\pm$0.8} & \underline{68.1}{\tiny$\pm$0.7} & \textbf{68.4}{\tiny$\pm$1.0} \\
\addlinespace
\multirow{2}{*}{OrganA}
 & 2\% & 83.6{\tiny$\pm$2.1} & 63.1{\tiny$\pm$0.6} & 72.9{\tiny$\pm$1.8} & 68.6{\tiny$\pm$2.2} & 84.9{\tiny$\pm$1.1} & 83.3{\tiny$\pm$1.5} & \underline{86.3}{\tiny$\pm$0.7} & \textbf{86.5}{\tiny$\pm$0.9} \\
 & 5\% & 89.8{\tiny$\pm$1.2} & 80.9{\tiny$\pm$0.8} & 84.6{\tiny$\pm$0.2} & 81.0{\tiny$\pm$2.0} & 90.5{\tiny$\pm$0.9} & \underline{90.7}{\tiny$\pm$0.6} & 90.4{\tiny$\pm$0.7} & \textbf{91.9}{\tiny$\pm$0.3} \\
\addlinespace
\multirow{2}{*}{Path}
 & 2\% & 77.5{\tiny$\pm$4.6} & 36.8{\tiny$\pm$1.0} & 53.5{\tiny$\pm$3.9} & 53.5{\tiny$\pm$5.7} & 77.0{\tiny$\pm$1.6} & 73.0{\tiny$\pm$3.8} & \underline{78.0}{\tiny$\pm$2.1} & \textbf{80.9}{\tiny$\pm$0.9} \\
 & 5\% & 84.0{\tiny$\pm$2.4} & 49.9{\tiny$\pm$2.4} & 58.9{\tiny$\pm$4.2} & 58.7{\tiny$\pm$3.5} & 82.5{\tiny$\pm$2.2} & 83.0{\tiny$\pm$1.6} & \underline{84.6}{\tiny$\pm$2.7} & \textbf{85.9}{\tiny$\pm$1.7} \\
\addlinespace
\multirow{2}{*}{Tissue}
 & 2\% & \underline{43.1}{\tiny$\pm$1.0} & 10.1{\tiny$\pm$0.4} & 39.3{\tiny$\pm$0.6} & 39.8{\tiny$\pm$0.6} & \textbf{44.7}{\tiny$\pm$1.0} & 35.6{\tiny$\pm$0.9} & 42.9{\tiny$\pm$0.9} & 42.6{\tiny$\pm$0.6} \\
 & 5\% & \underline{49.1}{\tiny$\pm$0.4} & 15.2{\tiny$\pm$0.2} & 43.6{\tiny$\pm$1.0} & 46.7{\tiny$\pm$1.0} & 48.1{\tiny$\pm$0.6} & 41.0{\tiny$\pm$1.1} & 48.3{\tiny$\pm$0.4} & \textbf{49.5}{\tiny$\pm$1.3} \\
\addlinespace
\multirow{2}{*}{Blood}
 & 2\% & \underline{83.2}{\tiny$\pm$1.5} & 48.3{\tiny$\pm$6.9} & 67.5{\tiny$\pm$3.3} & 75.9{\tiny$\pm$2.5} & 82.7{\tiny$\pm$1.6} & 78.8{\tiny$\pm$2.1} & 83.1{\tiny$\pm$3.0} & \textbf{84.5}{\tiny$\pm$1.7} \\
 & 5\% & 90.3{\tiny$\pm$0.5} & 72.5{\tiny$\pm$1.7} & 81.4{\tiny$\pm$2.1} & 86.6{\tiny$\pm$1.7} & 88.2{\tiny$\pm$2.7} & 89.0{\tiny$\pm$2.5} & \underline{92.3}{\tiny$\pm$0.6} & \textbf{93.4}{\tiny$\pm$0.7} \\
\bottomrule
\end{tabular}
\end{table}

\subsection{Analysis}

\noindent\textbf{Ablation study.}
Table~\ref{tab:ablation} summarizes ablations on kernel design.
Performance is stable across propagation depths: the differences between 1-hop, 2-hop, and 3-hop are small and inconsistent in direction on both PathMNIST and OrganAMNIST (at $k{=}5$ to amplify hop differences), suggesting that additional hops neither reliably help nor hurt.
Coverage in a $k$-NN graph is inherently local: selecting a sample already ``covers'' its direct neighbors, and propagation beyond two hops reaches samples that are better served by their own nearby representatives.
Replacing our polynomial kernel with the full heat kernel $\exp(-t\mathbf{L})$ yields comparable performance (Table~\ref{tab:ablation}).
Performance generally improves with graph density and stabilizes around $k{=}20$, though BloodMNIST at 2\% continues to benefit from denser graphs ($k{=}50$).
Very sparse graphs ($k{=}5$) underperform on BloodMNIST and OrganSMNIST at 2\%, suggesting that low budgets require sufficient neighborhood structure.
The sparse kernel is also necessary at scale: on TissueMNIST (165K samples), facility location on the full pairwise matrix exceeds GPU memory (102\,GB), while graph construction and selection complete in 89 seconds.

\begin{table}[t]
\caption{Ablation study. Balanced accuracy (\%) under different kernel configurations.}\label{tab:ablation}
\centering
\setlength{\tabcolsep}{3pt}
\fontsize{8}{9.5}\selectfont
\begin{tabular}{l cc cc}
\toprule
& \multicolumn{2}{c}{PathMNIST} & \multicolumn{2}{c}{OrganAMNIST} \\
\cmidrule(lr){2-3} \cmidrule(lr){4-5}
& 2\% & 5\% & 2\% & 5\% \\
\midrule
\multicolumn{5}{l}{\textit{Propagation depth}} \\[2pt]
\quad $\mathbf{A}$ (1-hop) & 79.4{\tiny$\pm$2.2} & 85.4{\tiny$\pm$2.3} & 86.7{\tiny$\pm$0.6} & 91.0{\tiny$\pm$0.6} \\
\quad $\mathbf{A}{+}\mathbf{A}^2$ (2-hop) & 81.0{\tiny$\pm$2.3} & 83.9{\tiny$\pm$1.4} & 87.1{\tiny$\pm$0.2} & 91.2{\tiny$\pm$0.4} \\
\quad $\mathbf{A}{+}\mathbf{A}^2{+}\mathbf{A}^3$ (3-hop) & 80.3{\tiny$\pm$0.8} & 84.5{\tiny$\pm$2.1} & 87.7{\tiny$\pm$0.3} & 90.6{\tiny$\pm$1.1} \\
\midrule
& \multicolumn{2}{c}{BloodMNIST} & \multicolumn{2}{c}{OrganSMNIST} \\
\cmidrule(lr){2-3} \cmidrule(lr){4-5}
& 2\% & 5\% & 2\% & 5\% \\
\midrule
\multicolumn{5}{l}{\textit{Kernel type}} \\[2pt]
\quad Polynomial ($\mathbf{A}{+}\mathbf{A}^2$) & 84.5{\tiny$\pm$1.7} & 93.4{\tiny$\pm$0.7} & 63.6{\tiny$\pm$0.7} & 68.4{\tiny$\pm$1.0} \\
\quad Heat kernel ($e^{-t\mathbf{L}}$) & 84.6{\tiny$\pm$1.5} & 93.2{\tiny$\pm$0.7} & 62.4{\tiny$\pm$0.6} & 68.0{\tiny$\pm$0.5} \\
\quad Coverage $\rho$ & \multicolumn{2}{c}{0.999} & \multicolumn{2}{c}{0.999} \\
\addlinespace
\multicolumn{5}{l}{\textit{Neighborhood size}} \\[2pt]
\quad $k = 5$ & 81.7{\tiny$\pm$2.4} & 92.5{\tiny$\pm$1.1} & 60.1{\tiny$\pm$1.1} & 68.1{\tiny$\pm$0.5} \\
\quad $k = 20$ & 81.9{\tiny$\pm$1.8} & 92.5{\tiny$\pm$1.2} & 63.6{\tiny$\pm$0.7} & 68.0{\tiny$\pm$0.5} \\
\quad $k = 50$ & 84.5{\tiny$\pm$1.7} & 93.4{\tiny$\pm$0.7} & 63.7{\tiny$\pm$0.7} & 68.4{\tiny$\pm$1.0} \\
\bottomrule
\end{tabular}
\end{table}

\begin{figure}[t]
\centering
\begin{minipage}[c]{0.32\linewidth}
\centering
\fontsize{8}{9.5}\selectfont
\captionof{table}{Global vs.\ per-class graph on DermaMNIST.}\label{tab:global}
\setlength{\tabcolsep}{3pt}
\begin{tabular}{l cc}
\toprule
Scope & 2\% & 5\% \\
\midrule
Per-class & 38.1{\tiny$\pm$1.3} & 48.9{\tiny$\pm$1.6} \\
Global & \textbf{43.8}{\tiny$\pm$2.6} & \textbf{52.3}{\tiny$\pm$1.9} \\
\addlinespace
$\Delta$ & +5.7 & +3.4 \\
\bottomrule
\end{tabular}
\end{minipage}
\hfill
\begin{minipage}[c]{0.65\linewidth}
\centering
\includegraphics[width=\linewidth]{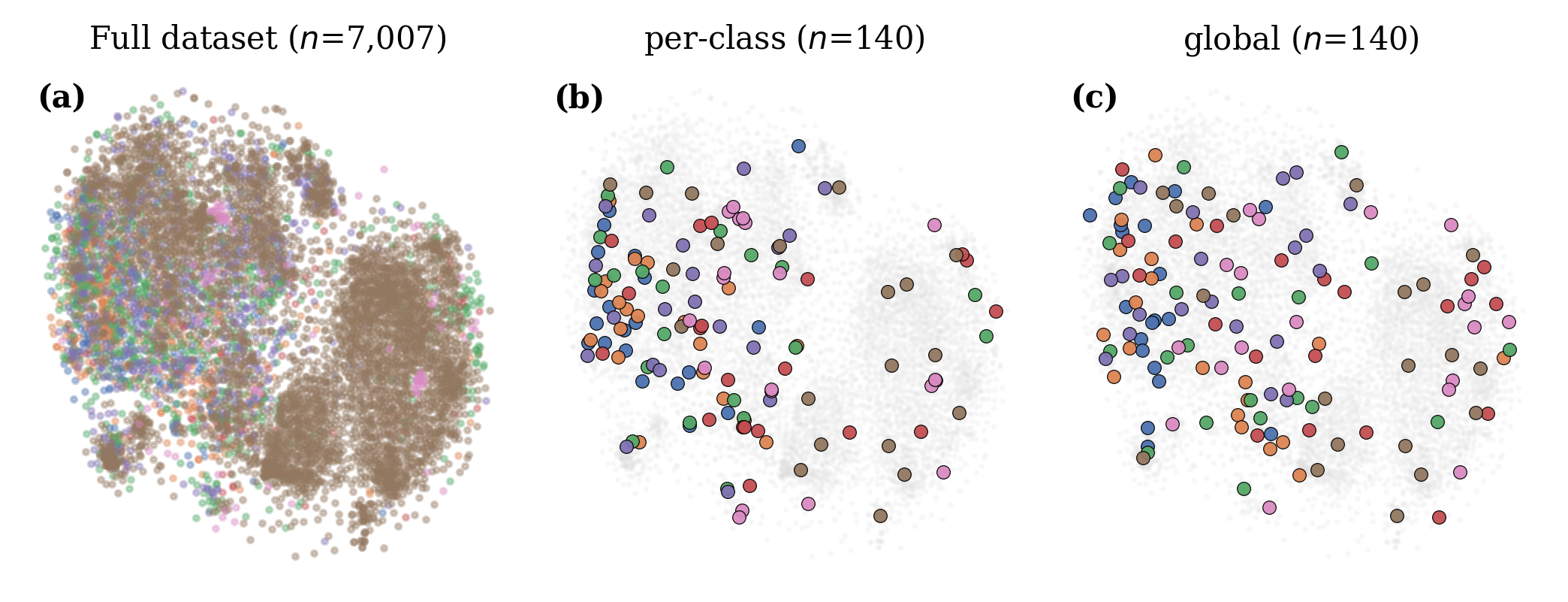}
\end{minipage}
\caption{Effect of global graph construction on DermaMNIST. (a)~Full dataset in t-SNE space; the majority class (brown) dominates. (b)~Per-class selection clusters within individual class regions. (c)~Global selection distributes samples into cross-class overlap zones.}\label{fig:global}
\end{figure}

\noindent\textbf{Global vs.\ per-class graph construction.}
Table~\ref{tab:global} and Fig.~\ref{fig:global} compare global and per-class graph construction on DermaMNIST, which exhibits severe class imbalance (58:1 ratio).
Global construction improves balanced accuracy by 5.7\,pp at 2\% budget.
The mechanism is visible in Fig.~\ref{fig:global}: per-class graphs for minority classes contain so few nodes that they are nearly complete, leaving little room for selection to improve coverage beyond random.
Global construction connects minority-class samples to the broader manifold, so the greedy algorithm selects representatives near decision boundaries rather than redundant intra-class neighbors.
Herding, which is inherently per-class, achieves 39.6\% on DermaMNIST at 2\%, closer to our per-class baseline (38.1\%) than to global construction (43.8\%), suggesting that methods without cross-class structure cannot exploit this advantage.
The gap narrows at 5\% ($+$3.4\,pp), consistent with minority classes having enough budget to cover their own regions at higher ratios.
The same pattern holds on BloodMNIST (84.5\% vs.\ 83.6\% at 2\%) and OrganSMNIST (63.7\% vs.\ 62.5\%).

\noindent\textbf{Training-dynamics methods.}
EL2N, forgetting, and EVA consistently underperform random selection on nearly every dataset and ratio (Table~\ref{tab:main}).
EL2N achieves 36.8\% on PathMNIST at 2\% compared to 77.5\% for random, and 10.1\% on TissueMNIST (near chance for 8 classes).
These methods select hard or ambiguous samples, which are informative for pruning large datasets but counterproductive at low budgets where the model needs representative prototypes, not boundary cases~\cite{sorscher2022beyond}.

\section{Conclusion}

We presented a one-shot data selection method for medical image classification that builds a $k$-NN graph on frozen foundation model embeddings and selects subsets via greedy facility location on a two-term coverage kernel.
The approach requires no model training before or during selection, and a single hyperparameter.

Three findings emerge from our experiments.
First, the value lies in graph structure, not propagation sophistication: the $k$-NN adjacency and its square capture nearly all selection-relevant information from the full spectral heat kernel, making eigendecomposition and decay coefficients unnecessary.
Second, global graph construction across classes consistently outperforms per-class construction, with the largest gains under class imbalance where minority-class graphs are too small for meaningful selection.
Third, training-dynamics methods are counterproductive at low budgets: they were designed for pruning large datasets, not selecting the 2--5\% of data where representative prototypes matter most.

The main limitation is dependence on embedding quality.
On TissueMNIST, where UNI embeddings poorly separate fluorescence microscopy classes, all selection methods converge, as graph structure cannot compensate for uninformative representations.
Extending this approach to other tasks and evaluating with domain-specific foundation models beyond pathology remain open directions.

% \begin{credits}
% \subsubsection{\ackname} Funding source, if any.
% \subsubsection{\discintname} The authors have no competing interests to declare that are relevant to the content of this article.
% \end{credits}


\bibliographystyle{splncs04}
\bibliography{references}

\end{document}